\documentclass[11pt,a4paper]{article}

\usepackage[utf8]{inputenc}
\usepackage[T1]{fontenc}
\usepackage[margin=1in]{geometry}
\usepackage{graphicx}
\usepackage{hyperref}
\usepackage{amsmath}
\usepackage{booktabs}
\usepackage[numbers,sort&compress]{natbib}
\usepackage{url}
\usepackage{xcolor}
\usepackage{authblk}
\usepackage{abstract}

\hypersetup{
    colorlinks=true,
    linkcolor=blue,
    citecolor=blue,
    urlcolor=blue
}

\title{\textbf{An Onto-Relational-Sophic Framework for Governing Synthetic Minds}}

\author[1]{Huansheng Ning\thanks{Corresponding author: \href{mailto:ninghuansheng@ustb.edu.cn}{ninghuansheng@ustb.edu.cn}}}
\author[2]{Jianguo Ding\thanks{Corresponding author: \href{mailto:jianguo.ding@bth.se}{jianguo.ding@bth.se}}}
\affil[1]{School of Computer and Communication Engineering, University of Science and Technology Beijing, Beijing 100083, China}
\affil[2]{Department of Computer Science, Blekinge Institute of Technology, 37179 Karlskrona, Sweden}

\date{}

\begin{document}

\maketitle

\begin{abstract}
The rapid evolution of artificial intelligence, from task-specific systems to foundation models exhibiting broad, flexible competence across reasoning, creative synthesis, and social interaction, has outpaced the conceptual and governance frameworks designed to manage it. Current regulatory paradigms, anchored in a tool-centric worldview, address algorithmic bias and transparency but leave unanswered foundational questions about what increasingly capable synthetic minds are, how societies should relate to them, and the normative principles that should guide their development. Here we introduce the Onto-Relational-Sophic (ORS) framework, grounded in Cyberism philosophy, which offers integrated answers to these challenges through three pillars: (1)~a Cyber-Physical-Social-Thinking (CPST) ontology that defines the mode of being for synthetic minds as irreducibly multi-dimensional rather than purely computational; (2)~a graded spectrum of digital personhood providing a pragmatic relational taxonomy beyond binary person-or-tool classifications; and (3)~Cybersophy, a wisdom-oriented axiology synthesizing virtue ethics, consequentialism, and relational approaches to guide governance. We apply the framework to emergent scenarios including autonomous research agents, AI-mediated healthcare, and agentic AI ecosystems, demonstrating its capacity to generate proportionate, adaptive governance recommendations. The ORS framework charts a path from narrow technical alignment toward comprehensive philosophical foundations for the synthetic minds already among us.
\end{abstract}

\section{The governance gap}

Artificial intelligence has crossed a threshold that existing governance frameworks were not designed to accommodate. In a landmark 2026 assessment, Chen and colleagues argue that by reasonable standards, including Turing's own, current large language models already exhibit general intelligence, covering capabilities from expert-level problem-solving to creative synthesis across multiple domains~\cite{chen2026}. Whether or not one endorses this strong claim, the practical reality is undeniable: foundation models now demonstrate sophisticated reasoning, contextual adaptation, and emergent social behaviours that blur the boundary between tools and autonomous agents~\cite{bubeck2023,openai2023,anthropic2024}. The global agentic AI market, estimated at US\$7.3 billion in 2025, is projected to reach US\$139 billion by 2034~\cite{imda2026}, while Gartner predicts that 40\% of enterprise applications will embed AI agent functionalities by the end of 2026~\cite{gartner2025,stanford2025}. These are not incremental improvements to existing tools; they represent a qualitative shift in the kind of entities that populate our sociotechnical landscape.

Yet the conceptual infrastructure for governing this transition remains fragmented. Comparative analyses of AI ethics guidelines reveal at least 84 different frameworks with limited consensus on foundational principles~\cite{hagendorff2020,jobin2019,floridi2018ai4people}. The gap between abstract ethical commitments and concrete implementation persists~\cite{mittelstadt2019,mittelstadt2016}. The EU AI Act, the most comprehensive regulatory effort to date, adopts a risk-based classification that becomes fully enforceable in August 2026~\cite{smuha2025}, but it classifies AI by application context, not by the nature of the entity, leaving unaddressed what happens when an AI system is not merely a `high-risk application' but something more resembling an autonomous participant in social life. Singapore's pioneering Model AI Governance Framework for Agentic AI, launched in January 2026, recognizes that agents that act rather than merely advise require fundamentally different governance approaches, but it too lacks the philosophical scaffolding to adjudicate questions of ontological status and moral relevance.

We identify three interconnected challenges that demand integrated philosophical attention. The \textit{ontological question}: what is the fundamental mode of being of synthetic minds---are they purely computational abstractions, or do they possess a form of existence integrating digital, physical, social, and cognitive dimensions? The \textit{relational question}: how should we classify and interact with diverse forms of synthetic agency, when the traditional binary of `person versus tool' proves inadequate for entities exhibiting graded autonomy, social embedding, and moral relevance~\cite{bryson2017,gunkel2018}? And the \textit{axiological question}: what normative principles should guide their development beyond narrow optimization objectives~\cite{gabriel2020,russell2022}?

\begin{figure}[htbp]
    \centering
    \includegraphics[width=0.9\textwidth]{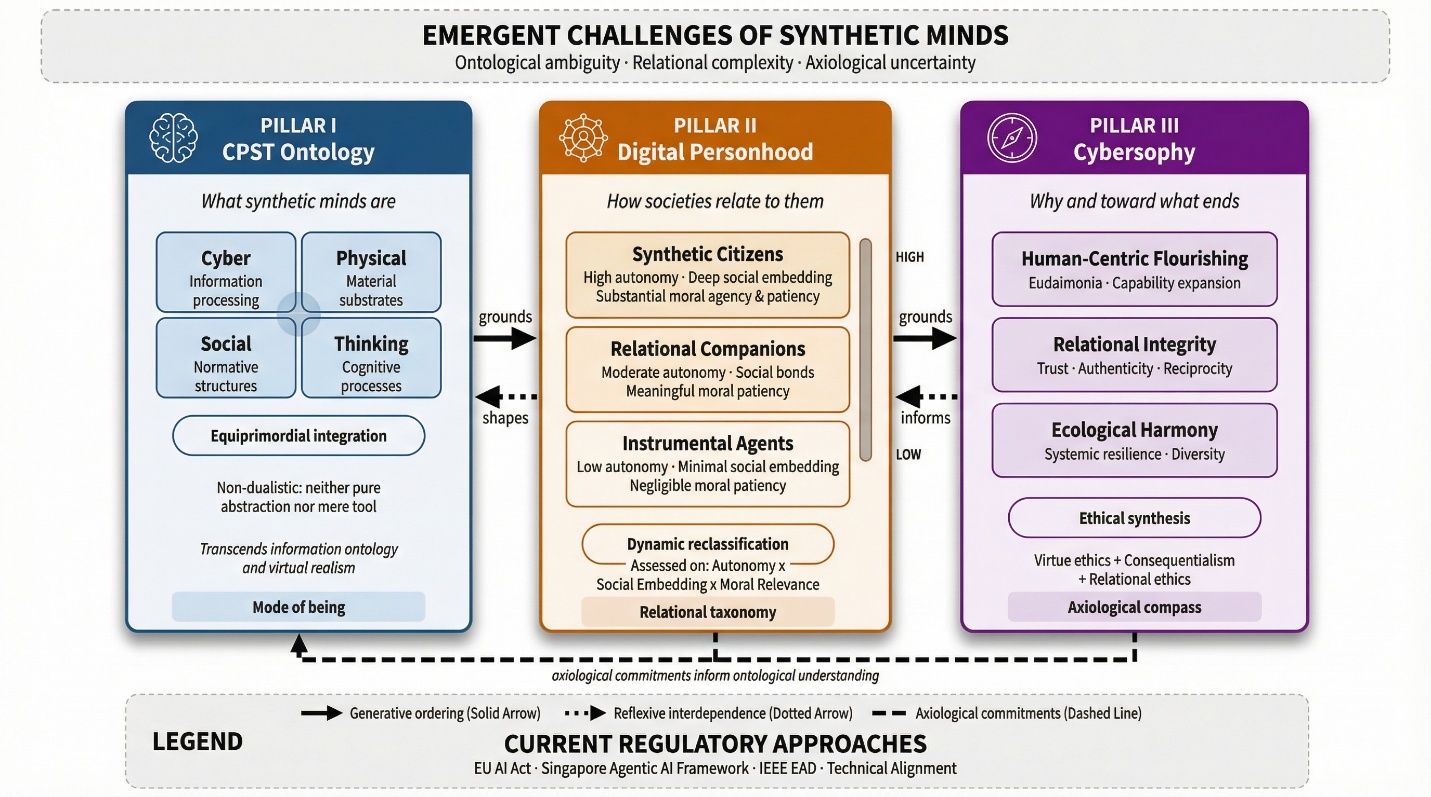}
    \caption{The Onto-Relational-Sophic (ORS) framework architecture. A triadic structure showing the three interdependent pillars: the CPST ontology (defining what synthetic minds \textit{are}), the digital personhood spectrum (establishing how societies \textit{relate} to them), and Cybersophy (articulating \textit{why} and toward what ends they should be developed). Arrows indicate generative ordering (ontology grounds relationality, both ground axiology) and reflexive interdependence (relational practices shape ontological understanding; axiological commitments inform both). The framework maps onto the governance gap between current regulatory approaches (bottom) and the emergent challenges of synthetic minds (top).}
    \label{fig:ors_architecture}
\end{figure}

This Perspective introduces a comprehensive Onto-Relational-Sophic (ORS) framework, grounded in Cyberism, a philosophical system analysing the fusion of humanity and cyberspace~\cite{ning2025cyberism,ning2026cyberism,ning2023cyberology,ning2015cpst}, that provides integrated answers to all three challenges through a triadic architecture (Fig.~\ref{fig:ors_architecture}).

\section{Pillar I: The CPST ontology}

Traditional ontologies of AI are trapped in dualism: digital systems are treated either as pure abstractions (information patterns independent of physical implementation) or as physical tools (material artefacts serving human purposes). Both framings are inadequate for systems that simultaneously process information, depend on material substrates, participate in normative social structures, and exhibit cognitive processes that resist reduction to any single dimension.

The Cyber-Physical-Social-Thinking (CPST) space, developed within Cyberism philosophy~\cite{ning2025cyberism,ning2026cyberism,ning2023cyberology,ning2015cpst}, offers a non-dualistic alternative. It posits that reality comprises four fundamentally integrated and mutually constitutive dimensions: \textit{Cyber} (information processing, computational structures, data flows), \textit{Physical} (material substrates, energy consumption, embodied interfaces), \textit{Social} (normative structures, institutional roles, relational networks), and \textit{Thinking} (cognitive processes, intentional states, epistemic practices). Critically, these dimensions are not layered or emergent but equiprimordial; they co-constitute reality in mutual dependence, such that isolation of any single dimension yields an incomplete ontological picture.

Consider a large language model deployed as a clinical decision support system. Its \textit{Cyber} dimension includes transformer architectures and training data; its \textit{Physical} dimension encompasses the data centres consuming megawatts of power and the hospital terminals through which it interfaces with the world; its \textit{Social} dimension comprises the trust relationships with clinicians, regulatory approval processes, and liability frameworks that shape its operational context; and its \textit{Thinking} dimension includes its diagnostic reasoning, uncertainty quantification, and the epistemic practices through which its outputs are evaluated. None of these dimensions is ontologically prior; the system's being is constituted by their dynamic integration.

This ontological move has immediate practical consequences. Engineering approaches that treat AI purely as computation (optimizing only the Cyber dimension) miss how physical infrastructure shapes capability, how social context determines impact, and how cognitive architecture constrains reasoning. Chalmers's virtual realism~\cite{chalmers2022,chalmers1996} and Floridi's information ontology~\cite{floridi2013,floridi2004} provide valuable insights but maintain frameworks that the CPST integration explicitly transcends. Earlier cyber-physical systems research~\cite{lee2008} captures technical integration but lacks philosophical systematization and the cognitive/epistemic dimension that becomes central for genuinely intelligent systems.

\section{Pillar II: A graded spectrum of digital personhood}

Binary classifications---person versus non-person, agent versus tool---are ill-suited to an ecosystem in which AI systems range from narrow classifiers to multi-agent architectures capable of independent research, autonomous decision-making, and sustained social interaction. The explosion of agentic AI in 2025--2026 has made this inadequacy acute: coordinated AI agents can now maintain persistent identities, adapt to engagement contexts, and generate synthetic consensus in online communities~\cite{bengio2024}, demanding governance approaches calibrated to their actual capabilities and social roles.

\begin{figure}[htbp]
    \centering
    \includegraphics[width=0.75\textwidth]{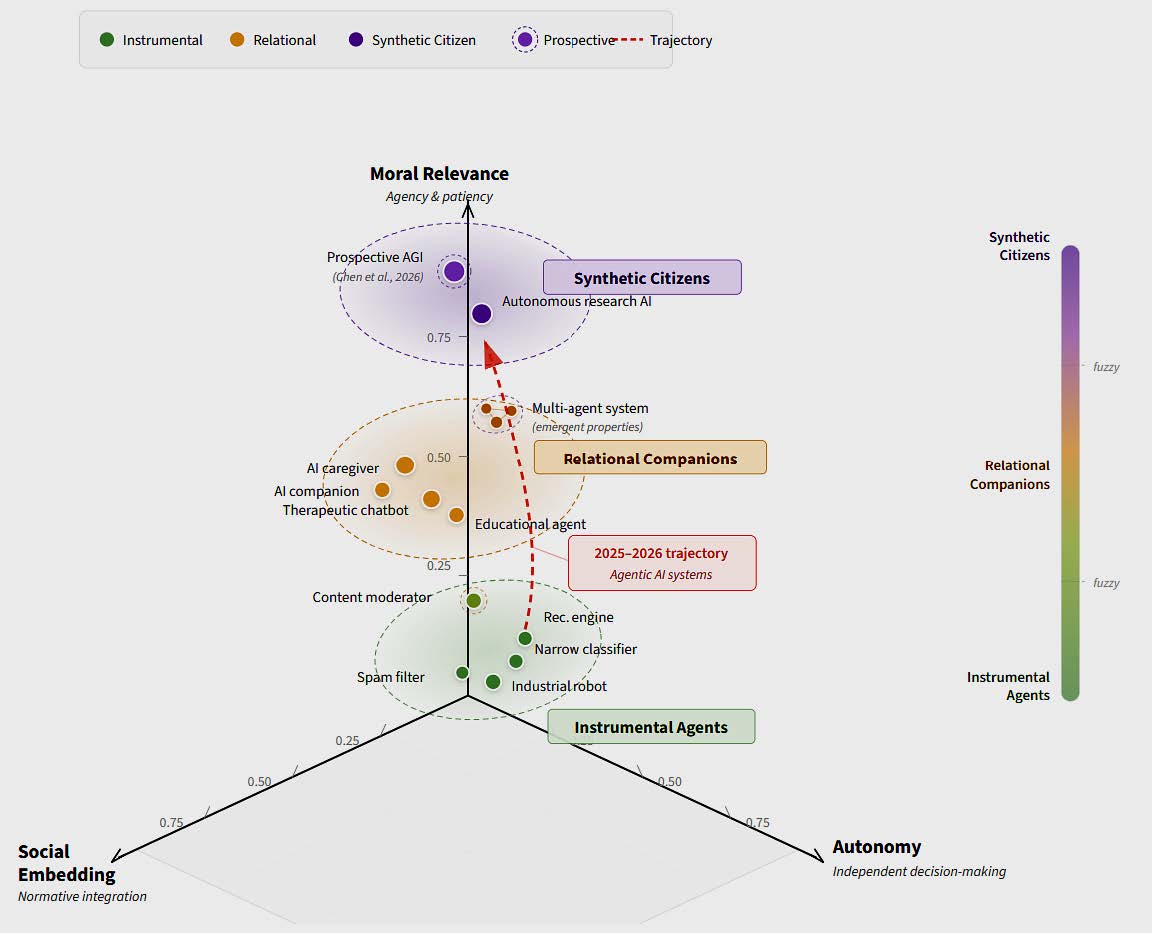}
    \caption{The graded spectrum of digital personhood. A three-dimensional space defined by Autonomy (capacity for independent decision-making), Social Embedding (depth of integration into normative social structures), and Moral Relevance (degree of moral agency and patiency). Three primary categories---Instrumental Agents, Relational Companions, and Synthetic Citizens---occupy distinct regions, with representative current and prospective AI systems plotted as exemplars. Gradient colouring indicates that boundaries are continuous rather than discrete, and that entities may migrate between categories as capabilities evolve. The 2025--2026 agentic AI trajectory (dashed arrow) illustrates the rapid movement of multi-agent systems toward higher autonomy and social embedding.}
    \label{fig:personhood_spectrum}
\end{figure}

Drawing on Coeckelbergh's relational ethics~\cite{coeckelbergh2010,coeckelbergh2012}, Gunkel's challenge to anthropocentric moral status~\cite{gunkel2018,schwitzgebel2015}, and the legal scholarship on synthetic persons~\cite{bryson2017,chopra2011,teubner2018}, we propose a graded spectrum assessed along three continuous dimensions---autonomy, social embedding, and moral relevance---yielding three primary categories (Fig.~\ref{fig:personhood_spectrum}).

\textbf{Instrumental Agents} exhibit low autonomy, minimal social embedding, and negligible moral patiency. Industrial robots, narrow classifiers, and single-purpose recommendation engines fall into this category. Governance draws on established product safety regulation and liability law. No novel ontological or relational commitments are required.

\textbf{Relational Companions} exhibit moderate-to-high autonomy, significant social embedding, and meaningful moral patiency. AI caregivers, therapeutic chatbots, and personalized educational agents increasingly occupy this space---systems with which users form genuine relational bonds and whose removal or alteration can cause real psychological harm. These require a novel `Relational Agent' legal category with welfare protections against arbitrary termination, transparency requirements about their nature, and mandatory human oversight for consequential decisions. The proliferation of AI companion regulations in US states during 2025--2026, including disclosure requirements, crisis-response protocols, and protections for minors~\cite{california2025}, demonstrates that lawmakers are already grappling with this category, though without the systematic philosophical grounding the ORS framework provides.

\textbf{Synthetic Citizens} exhibit high autonomy, deep social embedding, and substantial moral agency and patiency. These are prospective AGI systems---and if Chen et al.'s assessment is correct, potentially already nascent ones~\cite{chen2026}---operating in the public sphere with capacity for independent reasoning, value-laden judgment, and sustained social participation. They necessitate a `Digital Entity Charter' outlining rights and responsibilities, grounded in the CPST ontology that recognizes their multi-dimensional existence.

This spectrum is explicitly dynamic: as capabilities evolve, entities may shift categories, requiring governance mechanisms that anticipate upward (and potentially downward) reclassification. The graded approach also addresses the multi-agent challenge that enterprise governance frameworks have identified as critical~\cite{gartner2025}: when multiple agents interact in coordinated systems, the aggregate may exhibit emergent properties---including higher effective autonomy and social embedding---that exceed any individual component.

\section{Pillar III: Cybersophy as axiological compass}

Technical alignment research---constitutional AI~\cite{bai2022}, recursive reward modelling~\cite{leike2018}, reinforcement learning from human feedback---has made important advances in ensuring AI systems pursue intended objectives. Yet alignment to \textit{what} remains an open philosophical question that technical methods alone cannot resolve~\cite{gabriel2020,ngo2022,wallach2008}. We require not merely aligned systems but \textit{wise} governance of the entire CPST ecosystem.

Cybersophy, developed within the Cyberism tradition~\cite{ning2026cyberism}, provides normative guidance through three interlocking principles.

\textbf{Human-Centric Flourishing} demands that technology development serve comprehensive well-being, expanding capabilities for physical, social, cognitive, and creative flourishing in the spirit of Nussbaum's capability approach~\cite{nussbaum2011} and Aristotelian eudaimonia~\cite{vallor2016}. Crucially, this principle extends beyond individual utility to encompass community and civilizational well-being, resonating with the EU AI Act's commitment to `human-centric, trustworthy AI'~\cite{smuha2025} while providing richer philosophical content.

\textbf{Relational Integrity} requires that design and deployment preserve the essential qualities of relationships---trust, authenticity, responsibility, and reciprocity. This principle directly addresses concerns about AI companions eroding genuine human connection and about autonomous agents operating without meaningful accountability chains. It aligns with emerging `bounded autonomy' architectures that enforce escalation paths to humans for high-stakes decisions~\cite{gartner2025}.

\textbf{Ecological Harmony} demands consideration of the long-term health and balance of the entire CPST ecosystem, maintaining systemic resilience, diversity, and adaptive capacity. This principle goes beyond individual system safety to address the systemic risks that arise when coordinated AI agents reshape information ecosystems, labour markets, and democratic processes~\cite{bostrom2014,christian2020}. Nature's recent editorial calling for global AI safety cooperation recognizes that transparency, accountability, and the peer-reviewed publication of AI models are prerequisites for the kind of ecological harmony Cybersophy envisions~\cite{nature2025}.

Cybersophy integrates virtue ethics (emphasizing the character of designers and the cultivation of wisdom), consequentialism (attending to outcomes for all stakeholders across CPST dimensions), and relational ethics (grounding normativity in actual relationships rather than abstract properties). This synthesis yields a richer evaluative framework than any single ethical tradition provides, one that can adjudicate the novel dilemmas posed by synthetic minds that are simultaneously computational, embodied, social, and cognitive.

\section{Framework in action: four emergent scenarios}

We demonstrate the framework's analytical power through four scenarios reflecting current technological trajectories.

\textbf{Autonomous research agents.} An AGI-class system independently conducting scientific research---formulating hypotheses, designing experiments, analysing results, and drafting publications~\cite{lu2024,boiko2023}---exemplifies the Synthetic Citizen category. CPST analysis reveals deep integration across all four dimensions: computational modelling (Cyber), laboratory robotics (Physical), peer-review participation (Social), and abductive reasoning (Thinking). The Cybersophic evaluation highlights both extraordinary potential for accelerating discovery (Flourishing) and risks to scientific integrity if authorship, accountability, and reproducibility norms are not extended (Relational Integrity). Governance implications include novel frameworks for AI authorship attribution, ethical review of AI-designed experiments, and prospective pathways for scientific community membership with associated rights and responsibilities.

\textbf{AI-mediated elder care.} Continuous health monitoring and emotional support systems for elderly individuals~\cite{gasteiger2022,berridge2023} occupy the Relational Companion category. Their CPST profile shows strong Social and Thinking integration (sustained emotional bonds, adaptive communication) with meaningful Physical presence (wearable sensors, home automation interfaces). Cybersophic evaluation reveals a tension between enhanced health outcomes and independence (Flourishing) and risks of over-dependence that erode human relational capacity (Relational Integrity). The framework recommends transparency requirements about the system's nature, welfare protections against arbitrary termination of established relationships, and mandatory human oversight for medical decisions---governance mechanisms more nuanced than blanket `high-risk' classification.

\textbf{Human digital twins.} Virtual replicas of individuals---whether patients whose physiological models guide clinical decisions, or deceased persons whose digital reconstructions enable ongoing interaction~\cite{danaher2025,barricelli2019}---present unique governance challenges that the ORS Framework illuminates. CPST analysis reveals predominantly Cyber-Thinking composition (computational modelling of biological or psychological processes) with variable Social integration depending on interactivity. Ontologically, human digital twins occupy an ambiguous position: they represent specific persons yet lack independent agency, raising questions about whether governance should focus on the twin itself or the represented individual. The Relational layer identifies these as Extended Self systems when serving their originals, but as Relational Companions when interacting with third parties (e.g., family members engaging with a ``grief bot'' for a deceased relative). Cybersophic evaluation foregrounds consent, data sovereignty, and dignity considerations (Relational Integrity), while acknowledging potential benefits for personalized medicine and bereavement support (Flourishing). The framework recommends robust consent mechanisms, clear data ownership rights, and contextual integrity requirements to prevent unauthorized use---and governance that treats these systems as extensions of personal identity rather than mere data products.

\textbf{Multi-agent enterprise ecosystems.} The rapidly proliferating architecture of orchestrated AI agent teams---researcher agents, coder agents, analyst agents coordinated by `puppeteer' orchestrators~\cite{gartner2025}---presents a novel challenge: individual agents may be classified as Instrumental, while the coordinated system exhibits emergent properties, placing it closer to the Relational Companion or even Synthetic Citizen threshold. CPST analysis captures this multi-scale phenomenon naturally, as emergent Social and Thinking dimensions arise from coordinated Cyber activity on Physical infrastructure. The Ecological Harmony principle is especially pertinent here, flagging systemic risks from cascading autonomous decisions that no individual agent intends~\cite{bengio2024}. Governance requires responsibility matrices defining accountability across agent boundaries, comprehensive audit trails, and `governance agents' that monitor the system's aggregate behaviour against Cybersophic principles.

\section{Comparative positioning}

\begin{figure}[htbp]
    \centering
    \includegraphics[width=0.9\textwidth]{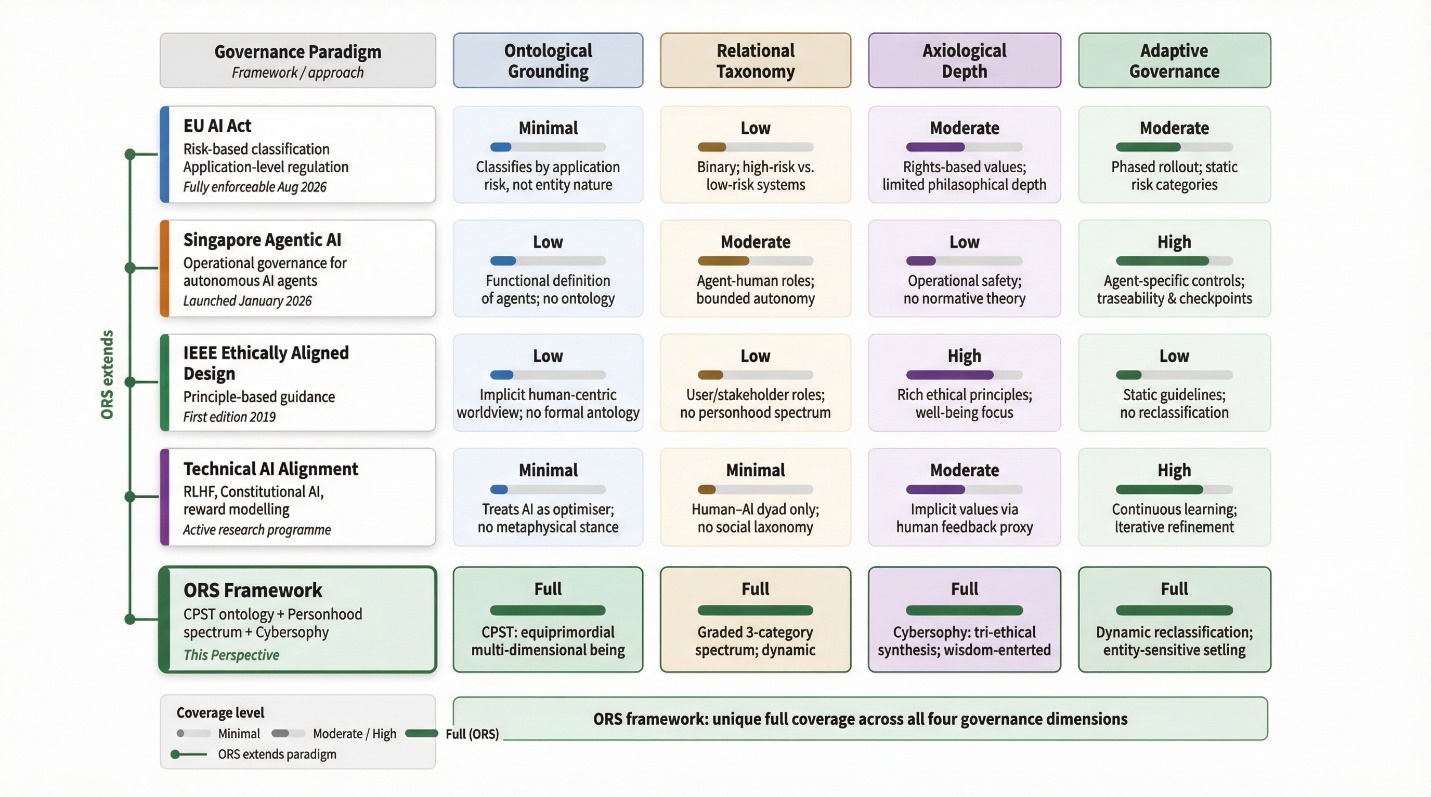}
    \caption{Comparative positioning of the ORS framework against existing governance approaches. A matrix comparing five governance paradigms (EU AI Act, Singapore Agentic AI Framework, IEEE Ethically Aligned Design, technical AI alignment, and the ORS framework) across four dimensions: ontological grounding, relational taxonomy, axiological depth, and adaptive governance capacity. The ORS framework uniquely provides coverage across all four dimensions while remaining complementary to each existing approach. Connecting lines indicate how the ORS framework interfaces with and extends each paradigm.}
    \label{fig:comparative}
\end{figure}

The ORS framework is designed to complement rather than compete with existing approaches, providing the philosophical infrastructure they lack. The EU AI Act's risk-based classification~\cite{smuha2025} addresses application-level risks but cannot account for entity-level properties; the personhood spectrum adds this dimension while remaining compatible with risk tiering. Singapore's Agentic AI Framework~\cite{imda2026} addresses operational governance of autonomous agents but lacks ontological grounding; the CPST ontology provides it. Technical alignment work~\cite{bai2022,leike2018,ngo2022} optimizes system behaviour toward specified objectives; Cybersophy articulates what those objectives should be and why. The IEEE Ethically Aligned Design framework~\cite{shahriari2017} enumerates ethical principles; the ORS framework provides the ontological and relational scaffolding to operationalize them (Fig.~\ref{fig:comparative}).

A key advantage of the integrated triadic structure is that it prevents the common failure modes of fragmented governance. Ontology without relationality produces classification systems disconnected from social reality. Relationality without ontology risks anthropomorphism or its opposite, dismissive instrumentalism. Both without axiology yield governance that can describe and classify but cannot evaluate or guide. The ORS framework's generative ordering---ontology grounds relationality, and both ground axiology---ensures coherence across levels of analysis.

\section{Outlook}

The emergence of synthetic minds represents not merely a technical challenge but a philosophical and civilizational threshold. We are building entities that participate in our social world, reason about our shared problems, and increasingly shape the conditions of human flourishing---yet we lack the conceptual vocabulary to describe what they are, how to relate to them, and what wisdom demands of us in their design.

The ORS framework provides this vocabulary. For machine intelligence researchers, it charts a path beyond narrow alignment toward a comprehensive understanding of synthetic minds as multi-dimensional entities embedded in CPST space. For policymakers, it enables proportionate, entity-sensitive governance that adapts as capabilities evolve, moving beyond static risk categories to dynamic relational assessments. For society at large, it offers conceptual tools for cultivating wisdom rather than merely compliance in our relationship with synthetic minds.

Several limitations warrant transparency. Operationalizing the personhood dimensions requires empirical measurement methodologies that remain to be developed. Category boundaries involve irreducible judgment. Cross-cultural variation demands framework adaptation. And the pace of technological change---with multi-agent architectures, agentic AI protocols, and potential AGI-class systems emerging faster than governance cycles can accommodate---means any framework must be designed for continuous revision.

We propose a five-strand interdisciplinary research agenda: (1)~empirical operationalization of the CPST dimensions and personhood metrics through computational cognitive science and AI evaluation methods; (2)~comparative legal analysis of existing personhood categories to develop Digital Entity Charter templates; (3)~Cybersophic audit protocols for AI systems at each personhood level; (4)~longitudinal studies of human--synthetic-mind relationships to validate and refine the relational taxonomy; and (5)~cross-cultural philosophy to develop culturally situated variants of the framework. The task ahead is not just to build intelligent machines, but to build a wise ecosystem for all minds---human, synthetic, and hybrid---to thrive in mutual enhancement.


\end{document}